\documentclass[letterpaper, 10 pt, conference]{ieeeconf}  

\IEEEoverridecommandlockouts                              

\usepackage{graphicx} 
\usepackage[font=small,labelfont=bf]{caption}
\usepackage{float}
\usepackage{amsmath} 
\usepackage{amsfonts,amssymb}
\usepackage[dvipsnames]{xcolor}
\usepackage{algorithm, algorithmic}
\usepackage{tabularx, booktabs, makecell, caption}
\usepackage{tabularray}
\usepackage{siunitx}
\usepackage{pifont}
\renewcommand{\baselinestretch}{0.99}
\usepackage{cite}
\usepackage[most]{tcolorbox}
\usepackage{lipsum} 
\tcbset{
  promptstyle/.style={
    colback=gray!10,    
    colframe=gray!70,   
    fonttitle=\bfseries,
    boxrule=0.5pt,
    arc=2mm,            
    left=2mm,
    right=2mm,
    top=1mm,
    bottom=1mm,
    enhanced,
    breakable           
  }
}

\usepackage[colorlinks=true, linkcolor=red, citecolor=green, urlcolor=cyan]{hyperref}
\usepackage{capt-of}

\makeatletter
\def\algbackskip{\hskip-\ALG@thistlm}
\makeatother

\title{\LARGE \bf
Vision-Language-Policy Model for Dynamic Robot Task Planning
}

\author{Jin Wang$^{1,2,*}$, Kim Tien Ly$^{1}$, Jacques Cloete$^{1}$, Jin Jin$^{1}$, Nikos Tsagarakis$^{2}$, Ioannis Havoutis$^{1}$ 
\thanks{†This work was supported by the European Union’s Horizon 2020 research and innovation programme, euROBIN EPUE034001, and Leonardo Joint Lab JL Leonardo ETCM058501; and the EPSRC project EP/Z531212/1. }
\thanks{$^{1}$Dynamic Robot Systems Group, Oxford Robotics Institute, University of Oxford, Oxford, U.K.}
\thanks{$^{2}$Humanoids and Human-Centered Mechatronics (HHCM), Istituto Italiano di Tecnologia, Genoa, Italy.}
\thanks{$^{*}$Corresponding author: {\tt\small wangshin.v@gmail.com}}
}

\IEEEaftertitletext{%
  \noindent\begin{minipage}{\textwidth}
    \centering
    \includegraphics[width=\textwidth]{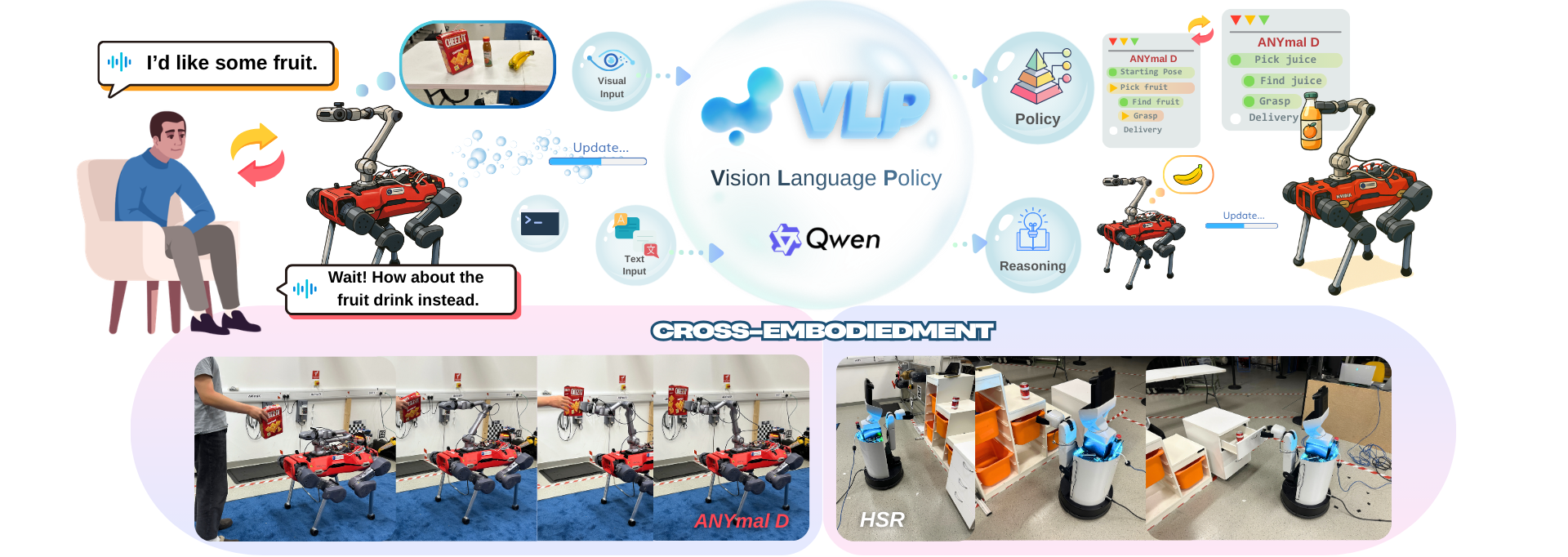}
    \captionof{figure}{We introduce the \textit{Vision-Language-Policy} model, designed for dynamic robot task planning. This innovative model surpasses state-of-the-art performance on semantic understanding, scene reasoning and policy updating through fine-tuning. Real-world experiments demonstrate that the model is applicable to various robotic task scenarios and can be deployed locally across different embodiments. }
    \label{fig:arch}
  \end{minipage}
  \vspace{0.75em}
}

\begin{document}

\maketitle
\thispagestyle{empty}
\pagestyle{empty}

\begin{abstract}
Bridging the gap between natural language commands and autonomous execution in unstructured environments remains an open challenge for mobile robots. This requires robots to perceive and reason over the current task scene through multiple modalities, and to plan their behaviors to achieve their intended goals. Traditional robotic task-planning approaches often struggle to bridge low-level execution with high-level task reasoning, and cannot dynamically update task strategies when instructions change during execution, which ultimately limits their versatility and adaptability to new tasks. In this work, we propose a language model-based framework for dynamic robot task planning. Our Vision-Language-Policy (VLP) model, based on a vision-language model fine-tuned on real-world data, can interpret semantic instructions and integrate reasoning over the current task scene to generate behavior policies that control the robot to accomplish the task. Moreover, it can dynamically adjust the task strategy in response to changes in the task, enabling flexible adaptation to evolving task requirements. Experiments conducted with different robots and a variety of real-world tasks show that the trained model can efficiently adapt to novel scenarios and dynamically update its policy, demonstrating strong planning autonomy and cross-embodiment generalization.
Videos: \url{https://robovlp.github.io/}
\end{abstract}
\vspace{1cm}

\section{INTRODUCTION}

Humans, through long-term learning and evolution, have developed the ability to understand semantic instructions and rapidly formulate plans based on their environment. This enables them to adaptively identify optimal solutions when facing complex physical interactions and dynamically changing solutions, and to execute actions step by step toward the desired goal. For example, when preparing a hamburger, a chef plans the sequence of steps and can modify the preparation strategy in real time according to the customer’s requests, such as ``no hot sauce'' or ``add extra bacon''.

Traditional robotic control and planning systems typically rely on precise physical modeling and accurate environment perception, which are largely applied within the task and motion planning (TAMP) paradigm \cite{garrett2020pddlstream,ly2024r,Srivastava2014}. Although such model-based approaches perform well in structured settings such as industrial assembly lines, their lack of semantic scene understanding and spatial reasoning capabilities makes it difficult for them to generate real-time, instruction-driven robot behavior plans in unstructured environments. 

In recent years, the rapid advancement of large language models (LLMs) and vision-language models (VLMs) has brought new breakthroughs to robotic learning, reasoning, and decision-making. These multimodal models demonstrate human-like capabilities in language understanding, spatial reasoning, and contextual perception. They can extract scene structures from images and videos, recognize object attributes and spatial relations, and infer the task requirements expressed through human language. This enables robots to autonomously generate control policies and behavioral plans for novel scenarios, and to perform tasks such as navigation and manipulation. Together, these developments represent an important step toward embodied intelligence and higher levels of robotic autonomy.

However, whilst integrating language models into robotic systems has outlined a promising avenue for enhancing robotic autonomy, significant challenges persist when applying such methods to real-world scenarios due to complex kinematics and physical interactions. Existing Vision-Language-Action (VLA) approaches attempt to generate robot actions directly from models, enabling robots to interpret task instructions and execute corresponding behaviors in diverse environments. Nevertheless, these methods typically rely on training with large-scale, high quality datasets, which are difficult to acquire, especially for mobile platforms such as legged manipulators where teleoperation data collection is challenging. Moreover, a substantial gap persists between the high-level semantic outputs of language models and the low-level motion control required for robotic execution, raising concerns about interpretability in behavior planning. Additionally, deploying such generic models efficiently and locally on robots remains an open problem.

To address these challenges, we introduce a deployable robot planning framework built on a \textit{Vision-Language-Policy} (VLP) model, which leverages semantic scene reasoning and spatial understanding to generate task strategies, thereby guiding autonomous motion and manipulation in the real world. Powered by fine-tuning a pre-trained VLM and using real-world interaction data, the proposed control system can interpret human instructions across diverse scenarios and, through coordinated calls to predefined action and perception APIs, produce executable control policies that bridge high-level planning with low-level actuation in real time. Furthermore, the model rapidly adapts to evolving command requirements during tasks, integrating prior working memory with state updates to dynamically refine plans. This enables mobile robots to flexibly adapt across diverse tasks and dynamic environments, achieving autonomous decision-making comparable to human reasoning.

The main contributions of this work are as follows:
\begin{itemize}
\item We introduce a planning framework for dynamic policy generation by integrating a grounded multimodal language model with behavior primitives and affordance perception modules, enabling the mobile robot to respond to evolving task requirements in real time.
\item We propose a local deployable VLP model trained on real-world physical interaction data. The model integrates semantic reasoning with spatial geometry and observations, to support online task reasoning and dynamic policy updates, effectively bridging high-level planning and low-level execution.
\item We validate the work across different mobile manipulators and diverse tasks to validate its effectiveness and generality, facilitating seamless deployment of lightweight models while providing interpretable hierarchical planning.
\end{itemize}


\section{Related Work}
Traditional task-planning frameworks focus on solving state-transition problems for deterministic, static, finite, and fully observable systems\cite{ghallab2004automated}. Optimization-based and sampling-based\cite{10705419, wensing2023optimization,10260625, jallet2022constrained, khazoom2023optimal} motion planners search for feasible trajectories in continuous spaces under system constraints, while behavior-tree-based methods\cite{colledanchise2018behavior, wang2024autonomous, cloete2025adaptive} achieve task-level control through reusable hierarchical behavior modules. Recent efforts have explored learning-based approaches to achieve human-level motion performance. Reinforcement learning\cite{9779429, deng2025learningrecoverdynamicreward, jin2023resilientleggedlocalnavigation} and imitation learning\cite{qin2022dexmv, 11202206}, combined with large-scale computational training, have enabled robots to acquire new skills through trial-and-error and human demonstrations respectively. However, as robotic systems rapidly transition from structured factory environments to unstructured physical-interaction scenarios, higher levels of autonomy are required, demanding task reasoning, semantic understanding, and the ability to dynamically adapt to changes during execution.

Grounding pre-trained foundation models such as LLMs and VLMs has become a highly promising direction for robotics research. Extensive prior studies have explored the use of language model-based planners for task reasoning\cite{ichter2022do, ren2023robotsaskhelpuncertainty, song2023llmplanner, wang2024hypermotion, ly2024inteliplan}, code generation\cite{liang2022code, singh2022progpromptgeneratingsituatedrobot}, scene understanding\cite{wang2025intention, 11128775,jiang2026learning} and failure detection\cite{10679904,liu2023reflect, mei2024replanvlmreplanningrobotictasks, wang2024autonomous}. By integrating multi-modal inputs such as vision and audio, perception and motion can be directly associated with semantic instructions. However, most of these approaches rely on calling model APIs, making real-time, onboard inference infeasible. Moreover, they are constrained by robot-specific configurations, limiting their generalization across different embodiments. Meanwhile, VLA models\cite{kim2024openvla,intelligence2025pi05,li2024robonurse,brohan2023rt} are employed to generate robotic action sequences based on multimodal inputs, thereby addressing generalized manipulation tasks. Nevertheless, existing VLA approaches adopt a fully end-to-end paradigm, which is inherently non-interpretable and requires large-scale datasets for training. This often leads to high computational cost and susceptibility to overfitting, risking poor scene generalization and ultimately posing safety concerns in real-world settings. 


In contrast, our approach departs from end-to-end methods by explicitly factoring decision making into interpretable, structured policies. The proposed VLP model extracts object-centric interaction affordances under diverse scenes and performs instruction-driven reasoning to synthesize online, executable task policies, enabling dynamic re-planning and incremental policy updates when execution conditions change. Benefiting from post-training on real-world interaction data and efficient on-device deployment, the model rapidly parses semantic inputs and maps them to primitive-level constraints and decisions, making the overall pipeline transparent, verifiable, and easier to debug than black-box end-to-end learning. Real-world experiments across multiple mobile manipulators further show that, because the policy is generated in a platform-agnostic primitive space rather than robot-specific low-level commands, the method generalizes to multiple everyday tasks and different embodiments without additional retraining, while retaining the flexibility to adapt its strategy online.


\begin{figure*}
 \centerline{\includegraphics[width=14cm]{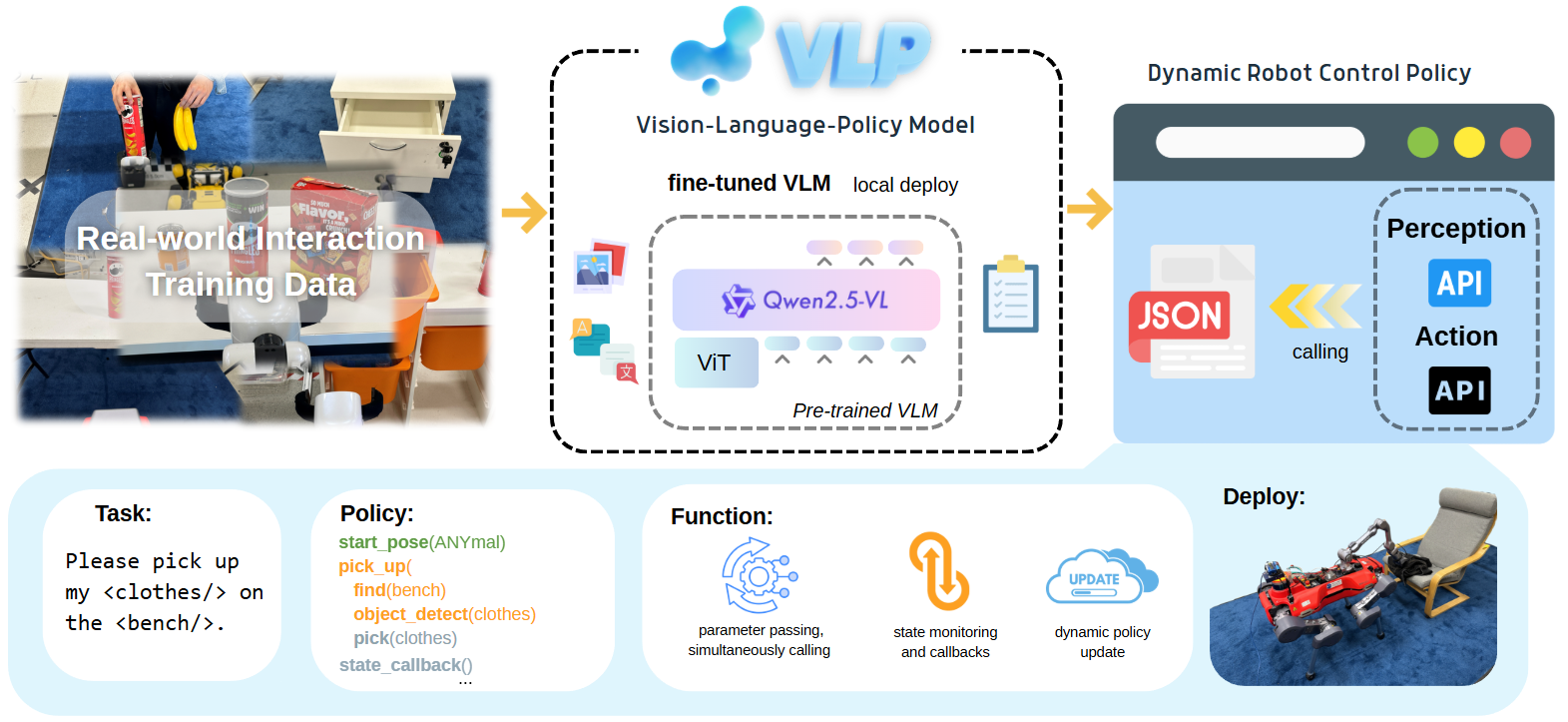}}
\caption{\textbf{System Overview.} Stage 1 performs post-training of the VLM using real-world interaction data consisting of images, task instructions, and corresponding policies. Stage 2 deploys the VLP model locally and generates structured policies based on semantic input to achieve real-time robot control and self-updating.}
\label{framework}   
\end{figure*}

\section{Methodology}

\subsection{Task Formulation}

We design a two-stage pipeline including model training and deployment for the proposed VLP model, as shown in Algorithm \ref{alg:vlp}. In the first stage, a pre-trained vision-language model $Q_v$ is fine-tuned using a real-world robot interaction dataset $\mathcal{D}$ via Low-Rank Adaptation (LoRA)\cite{hu2021loralowrankadaptationlarge}, resulting in a policy model $Q_p$ that is capable of task reasoning, spatial scene understanding and policy generation. During deployment stage, the robot operates continuously and receives a semantic instruction $I_t$, while a task memory module $\mathcal{M}$ is initialized to store contextual information. At each timestep, the system acquires the current visual observation $o_t$ and robot state $s_t$, and queries $Q_p$ together with the robot's behavior primitives $\mathcal{A}$ and memory $\mathcal{M}$ to produce a hierarchical policy $P_t$. Each behavior in $P_t$ is executed sequentially, with perception and state feedback updated after execution. If the task is completed, the loop terminates; otherwise, when either a new instruction is received or the task state changes (i.e., {\it strategic trigger}), both the instruction $I_t$ and memory $\mathcal{M}$ are updated, and $P_t$ is regenerated accordingly. This enables real-time adaptation to evolving requirements, effectively bridging high-level semantic reasoning with low-level robotic control in an interpretable and reactive manner.


\begin{algorithm}[t]
\caption{VLP Model Training and Deploy}
\label{alg:vlp}
\begin{algorithmic}[1]
\REQUIRE Pre-trained VLM $Q_v$, robot behavior library $\mathcal{A}$

\STATE \textbf{// Step 1: Model training}
\STATE Collect real-world robot interaction dataset $\mathcal{D}$
\STATE $Q_p \leftarrow \text{LoRA\_FineTune}(Q_v, \mathcal{D})$
\STATE \textbf{// Step 2: VLP Deployment}
\STATE Initialize task memory $\mathcal{M}$
\WHILE{$system\_is\_active$}
    \STATE Receive task instruction $I_t$
    \WHILE{$task\_not\_done$}
        \STATE Acquire visual observation $o_t$ and robot state $s_t$
        \STATE $P_t \leftarrow Q_p(I_t, o_t, s_t, \mathcal{A}, \mathcal{M})$
        \FOR{each behavior $a$ in $P_t$}
            \STATE Execute($a$)
            \STATE Update $(o_t, s_t)$
            \IF{$task\_completed$}
                \STATE \textbf{break}
            \ENDIF
            \IF{$new\_instruction$ \textbf{OR} $task\_state\_changed$}
                \STATE Update instruction $I_t$ if required
                \STATE Update memory $\mathcal{M}$
                \STATE \textbf{break} \COMMENT{policy update}
            \ENDIF
        \ENDFOR
    \ENDWHILE
\ENDWHILE

\end{algorithmic}
\end{algorithm}

\subsection{System Framework}

\begin{figure*}
 \centerline{\includegraphics[width=16cm]{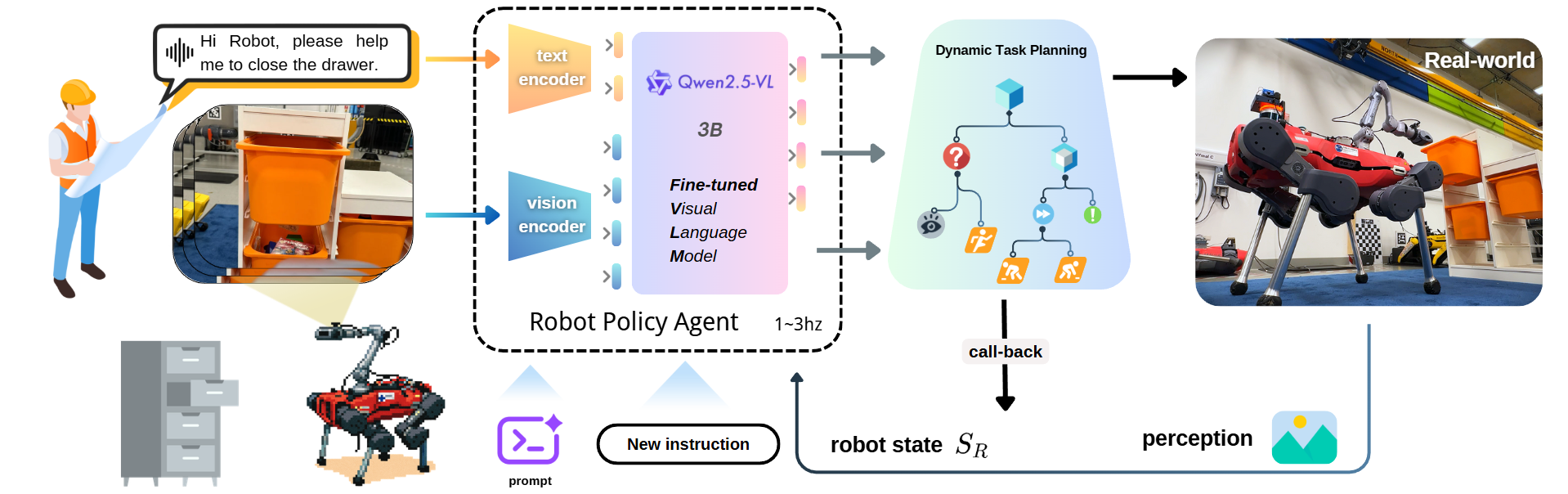}}
\caption{\textit{Vision-Language-Policy} model internal structure and real-world task scenario.}
\label{VLP}   
\end{figure*}

To address the problem of autonomous motion and task planning on complex robotic platforms (e.g., mobile manipulators and legged robotic systems) while mapping high-level semantic instructions to low-level control and execution, we propose a language-model based planning and control framework, as illustrated in Fig. \ref{framework}. We first conduct real-world interactive demonstrations on physical robots to collect training data comprising of task instructions, scene images, and the corresponding robot action strategies. This dataset is then used to fine-tune a pre-trained VLM, enabling the model to produce structured policies while preserving its inherent general reasoning capability. The resulting VLP model is deployed locally on the robot for real-time inference. To enable policy generation that effectively links semantic instructions with robot execution, we predefine a library of behavioral primitives, which includes both action and perception modules. These predefined primitives serve as fundamental motion and functional units and can be directly invoked by the generated policy. For each new task, the VLP model receives the instruction from the operator and reasons over the perceived physical scene to generate a policy that fulfills the task objective. During policy execution, the system continuously monitors the state and updates relevant parameters (e.g., execution status of action primitives and poses of target objects). If a new instruction is received mid-task, execution is temporarily paused, and the current task state, robot status, and execution history are retrieved and passed to the VLP together with the new instruction. The model then recomputes and updates the policy in real time, enabling the robot to autonomously replan and adapt to evolving task requirements during operation.

\subsection{Vision-Language-Policy Model}

The structure of the proposed VLP model and its deployment in the task environment are illustrated in Fig. \ref{VLP}. The VLP model is obtained by fine-tuning the pre-trained Qwen2.5-VL model\cite{bai2025qwen25vl}. Specifically, the model uses Qwen2.5 LLM\cite{qwen2025qwen25} as its backbone and employs a Vision Transformer (ViT) as the visual encoder. After fine-tuning, the LM decoder outputs hierarchical JSON-formatted policies, which are used to perform real-time planning of robot behaviors.

\begin{figure}
 \centerline{\includegraphics[width=6.5cm]{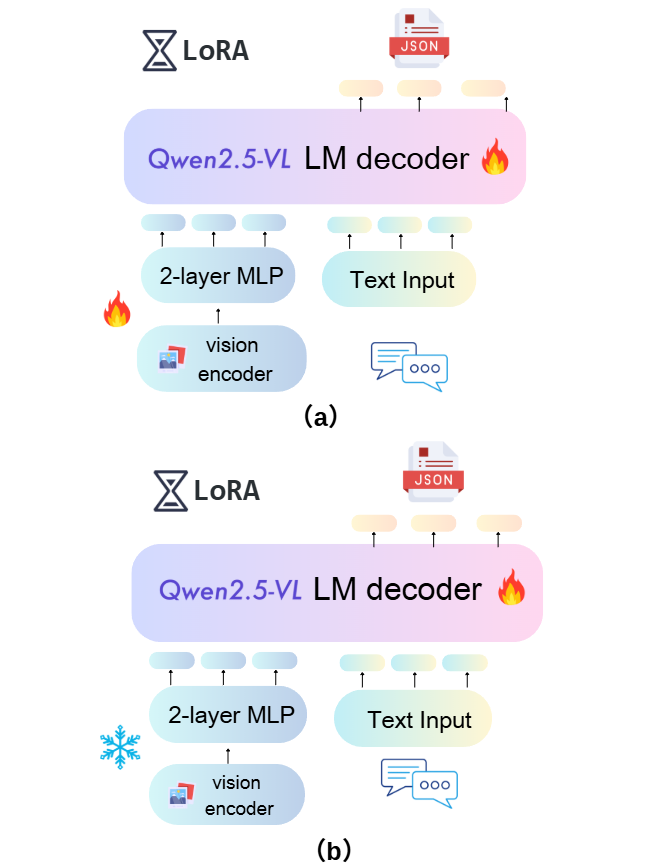}}
\caption{Comparison of different model fine-tuning strategies. \textbf{(a)}  Vision+Decoder LoRA: LoRA is applied to both the vision encoder and LM decoder. \textbf{(b)} Decoder-only LoRA: The vision encoder is frozen, and LoRA is applied solely to the LM decoder.}
\label{model_training}   
\end{figure}

\subsubsection{Model training}
In order to generate task-specific structured policies while preserving its inherent general understanding capability, we construct a post-training dataset by simulating human, robot, and environment interactions in real-world settings. It includes 15 different task environments and layouts, 12 different types of mobile and manipulation tasks, and involves interactions with various targets and objects. For each trial, the robot’s visual observations were captured through onboard cameras, and corresponding task policies were manually annotated based on the observed states and given task instructions.  We expanded the dataset to approximately 1,000 samples via multi-view acquisition from different robot embodiments and by extracting keyframes from demonstration videos. To retain the model’s original general understanding ability while enabling it to generate task-aligned policies in a structured JSON format, we implemented and compared two post-training strategies (as shown in Fig. \ref{model_training}) and fine-tuned the Qwen2.5-VL model. 

\textbf{Vision+Decoder LoRA}: LoRA adapters were applied to both the last 6 layers of the visual encoder and the LM decoder. A smaller rank (r = 8) was used for the visual side to mitigate overfitting while maintaining r = 16 on the decoder. This strategy aims to refine visual perception to better align with robot-centric viewpoints. \textbf{Decoder-only LoRA}: The visual encoder was fully frozen, and LoRA adapters were applied only to the language decoder. This configuration allows the model to adapt its reasoning and policy generation capabilities without altering the pre-trained visual representation. LoRA rank was set to r = 16, with a learning rate of 1e-4.

Both models were trained using a causal language modeling objective, with the loss computed only over the JSON policy generation segment, and the training was conducted for 2 epochs. Comparisons between the fine-tuned models and the benchmark baseline are provided in Section \ref{sec:experiments}.

\subsubsection{Policy generation}

The task policy generated by the VLP model is a JSON-formatted structure containing a sequence of action and perception behaviors. After being converted into an executable Python script, it can be directly used to control the robot’s motion. The design of the policy structure is inspired by behavior trees commonly used in robotic planning \cite{iovino2022survey}. While retaining their hierarchical representation and node-based logic, the structure is simplified into a top-down sequential execution scheme that supports real-time propagation and feedback of execution states and parameters. This lightweight, structured design not only ensures interpretability of action planning, but also enables dynamic policy updates, allowing timely adaptation of robot behaviors in response to new instructions or task requirements during execution.

\subsubsection{Motion primitives and visual perception}

To bridge policy planning with low-level robot execution, we predefined a set of action and perception behavior primitives that encode the robot’s intrinsic capabilities and can be directly invoked by the generated policies. This design enables the model to decompose complex tasks into sequences of skills, thereby enhancing the interpretability of each step in the task process and reducing the gap between high-level planning and low-level control.

Action primitives control robotic whole-body movements, such as basic locomotion and manipulation. Inspired by common object manipulation in domestic environments, we designed several robot actions, including \texttt{grasp}, \texttt{lift}, \texttt{place}, \texttt{handover}, etc. Each action is encapsulated as an independent API and accepts the Cartesian pose of the target as input. Additionally, to ensure smooth transitions and continuity throughout the execution, we define a set of preparatory and intermediate action poses, such as \texttt{homing}, \texttt{wake\_up}, and \texttt{start\_pose}, which support pre-operation alignment and motion transitions between consecutive actions.

Perception primitives are built upon the robot’s onboard sensing modules and are used to extract both object-related and self-state information. In addition, we leverage the model’s visual object grounding to perform spatial reasoning and affordance extraction. Combined with depth images, this enables generic object localization and grasp point generation, which are encapsulated as callable APIs and provided as perception primitives for policy execution.

\begin{tcolorbox}[promptstyle, fontupper=\ttfamily\small, title=Partial prompts for visual object grounding:]
You are a robotic perceptor. \\
Given the current scene \textcolor{red}{\{image\}} and task \textcolor{red}{\{instruction\}}, identify the target \textcolor{red}{<object>} specified in the \textcolor{red}{\{instruction\}}. 
Then, considering the next \textcolor{red}{<action>} defined in the policy, infer an appropriate manipulation \textcolor{red}{<point>} for the \textcolor{red}{<object>} that enables successful execution of the \textcolor{red}{<action>}.

\end{tcolorbox}


\begin{figure}
 \centerline{\includegraphics[width=8cm]{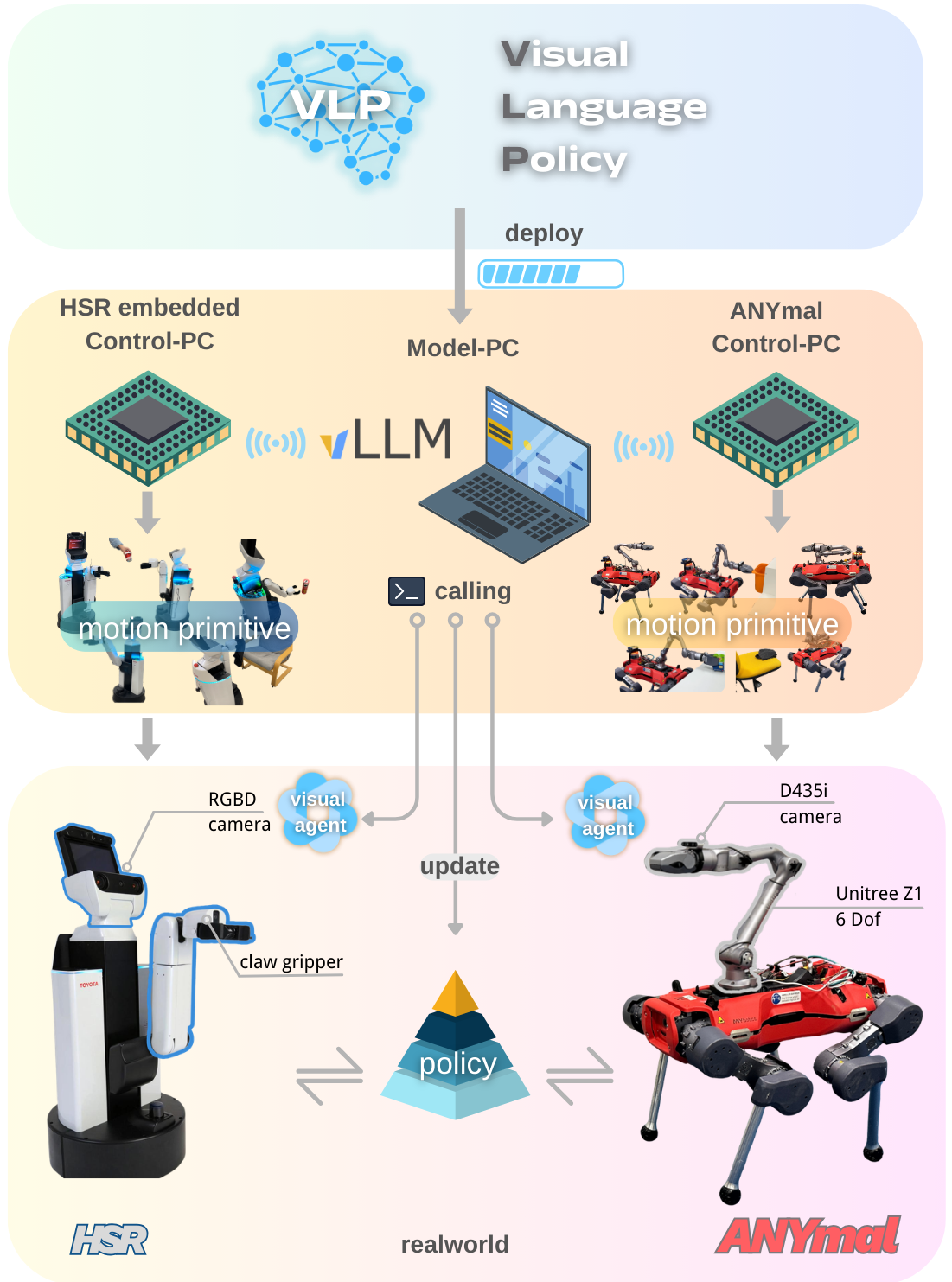}}
\caption{Real-world system setup for VLP model deployment.}
\label{setup}   
\end{figure}

\begin{table*}[t]
\caption{Comparison of Vision-Language-Policy model with existing robotic task planning and control methods.}
\label{tab:vs_baselines}
\centering
\begin{tblr}{
  cells = {c},
  colspec = {p{2.4cm} p{1.2cm} p{1.4cm} p{1.6cm} p{1.6cm} p{1.6cm} p{1.6cm} p{1.6cm}},
  cell{1}{1} = {r=2}{},
  cell{1}{2} = {c=7}{},
  hline{1,3,9} = {-}{0.16em},
  hline{3,8} = {-}{0.08em},
  hline{2} = {2-8}{},
  column{2-8} = {m, colsep=4pt},
}
\textbf{Method} & \textbf{Abilities} & & & & & & \\
   & \small{Autonomy} & \small{Language Feedback} & \small{Task Decomposition} & \small{Reactive Memory} & \small{Legged Manipulation} & \small{Locally Deploy} & \small{Online Replan} \\
BT-Planner     & \small{\textit{low}}   & \textcolor{red}{\ding{55}} & \textcolor{green!60!black}{\ding{52}} & \textcolor{red}{\ding{55}} & \textcolor{green!60!black}{\ding{52}} & \textcolor{green!60!black}{\ding{52}} & \textcolor{red}{\ding{55}} \\
LLM-BT         & \small{\textit{high}}  & \textcolor{green!60!black}{\ding{52}} & \textcolor{green!60!black}{\ding{52}} & \textcolor{red}{\ding{55}} & \textcolor{green!60!black}{\ding{52}} & \textcolor{red}{\ding{55}} & \textcolor{red}{\ding{55}} \\
OpenVLA        & \small{\textit{high}}  & \textcolor{red}{\ding{55}} & \textcolor{red}{\ding{55}} & \textcolor{green!60!black}{\ding{52}} & \textcolor{red}{\ding{55}} & \textcolor{green!60!black}{\ding{52}} & \textcolor{green!60!black}{\ding{52}} \\
$\pi_{0.5}$    & \small{\textit{high}}  & \textcolor{red}{\ding{55}} & \textcolor{red}{\ding{55}} & \textcolor{green!60!black}{\ding{52}} & \textcolor{red}{\ding{55}} & \textcolor{green!60!black}{\ding{52}} & \textcolor{green!60!black}{\ding{52}} \\
ReplanVLM      & \small{\textit{high}}  & \textcolor{green!60!black}{\ding{52}} & \textcolor{green!60!black}{\ding{52}} & \textcolor{red}{\ding{55}} & \textcolor{red}{\ding{55}} & \textcolor{red}{\ding{55}} & \textcolor{green!60!black}{\ding{52}} \\
\textbf{VLP (ours)} & \small{\textit{high}}  & \textcolor{green!60!black}{\ding{52}} & \textcolor{green!60!black}{\ding{52}} & \textcolor{green!60!black}{\ding{52}} & \textcolor{green!60!black}{\ding{52}} & \textcolor{green!60!black}{\ding{52}} & \textcolor{green!60!black}{\ding{52}} \\
\end{tblr}
\end{table*}

\section{Experiment and Evaluation}
\label{sec:experiments}

We performed a series of real-world experiments to validate the VLP-based dynamic task planning system, instructing real robots to complete various everyday manipulation tasks. This section provides detailed descriptions of the experimental setup and design, followed by an analysis and discussion of the results.

\subsection{Experiment setup}

We fine-tuned the \texttt{Qwen2.5-VL-3B-Instruct} model\cite{bai2025qwen25vl} using LoRA under two different settings: Decoder-only LoRA and Vision+Decoder LoRA. Each configuration was trained for 2 epochs using mixed precision (BF16). All training was conducted on a workstation equipped with an \texttt{NVIDIA RTX~6000~Ada} GPU (48\,GB VRAM) and 128\,GB RAM.

After training, the model was deployed locally using \texttt{vLLM}\cite{kwon2023efficient} on the \textit{Model-PC}, equipped with an \texttt{NVIDIA RTX~5080} GPU (16\,GB VRAM) and 32\,GB RAM. The deployment supports continuous batching and parallel multi-request inference, enabling efficient multi-threaded execution across different tasks. Upon receiving a task instruction, the robot's \textit{onboard PC} sends an HTTP request to the model, which typically generates a task policy with sub-second latency, sufficient for real-time planning and dynamic policy updates. Once the policy is running, predefined motion primitives are executed to control the robot at the low level. Additionally, perception primitives can internally and asynchronously invoke the model via API requests to perform specific perception tasks.

We validated the cross-embodiment capability of the proposed model on two different robotic platforms:

\begin{itemize}
    \item \textbf{ANYmal~D + Unitree~Z1 Arm}: A $12+6$ DoF platform consisting of a quadruped robot base and a lightweight manipulator featuring a claw gripper and a wrist-mounted \texttt{Intel RealSense D435i} depth camera.
    
    \item \textbf{Toyota Human Support Robot (HSR)}: A mobile robot with a single-arm configuration equipped with a 2-finger parallel gripper. The robot has an adjustable base height, and features a total of $11$ DoF. The head is equipped with an \texttt{Asus Xtion PRO Live} stereo camera.
\end{itemize}

The experimental setup is illustrated in Fig. \ref{setup}. All experimental scenes were set up in a laboratory environment using everyday household objects.

\subsection{Autonomous Task Planning Evaluation}

\begin{figure}
 \centerline{\includegraphics[width=8cm]{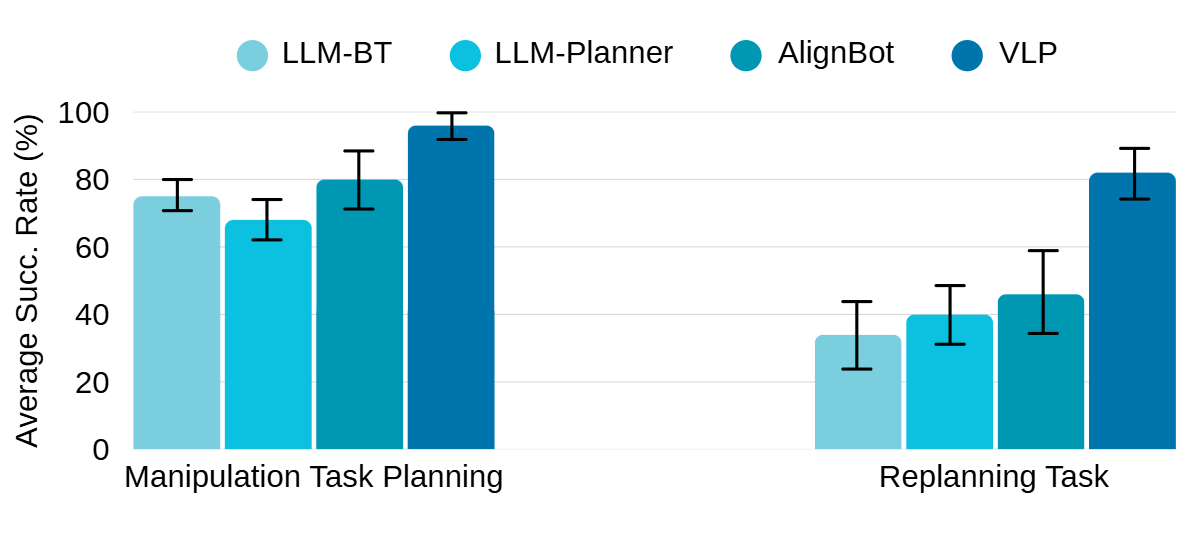}}
\caption{Comparing average success rates for task planning with existing methods.}
\label{comparison}   
\end{figure}

\subsubsection{Comparison with other methods}
We conducted a comparative analysis of the proposed framework against several existing approaches using a combination of qualitative and quantitative evaluations from multiple perspectives. First, we qualitatively compared the VLP model with methods\cite{wang2024autonomous,kim2024openvla,intelligence2025pi05,mei2024replanvlmreplanningrobotictasks} based on LLMs, VLMs, and VLA models across different capability dimensions. The results in Table \ref{tab:vs_baselines} show that our method outperforms VLA-based approaches in terms of task decomposition and planning interpretability. By leveraging local inference capability and online re-planning, it demonstrates superior adaptability and responsiveness compared to other language-model-based methods.

As shown in Fig. \ref{comparison}, the quantitative evaluation of planning success rates for everyday manipulation tasks and re-planning scenarios further demonstrates that the VLP exhibits superior planning capability across various task settings compared to other models\cite{wang2024autonomous, song2023llmplanner, 11128775}. In particular, when re-planning is required due to new command input, the incorporation of visual modality enables significantly higher success rates than language-model-based planning approaches.

\begin{figure*}
     \centerline{\includegraphics[width=16cm]{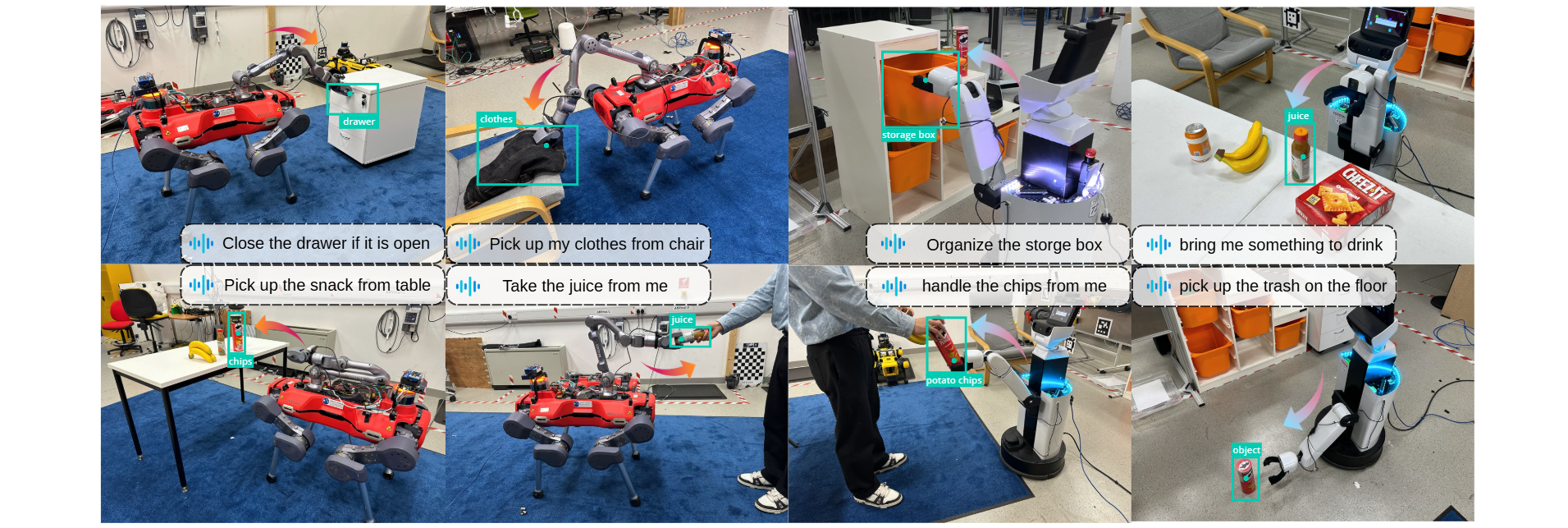}}
    \caption{Real-world experiments across various natural language tasks using the ANYmal and HSR robots, with demonstrations of model's capability in visual object grounding.}
    \label{realworld}
\end{figure*}

\subsubsection{Multi-task scenario evaluation}

\begin{table}[t]
\caption{Model vs. Task performance with plan and success rates for each task.}
\label{tab:models}
\centering
\begin{tblr}{
  cells = {c},
  colspec = {p{1.8cm} *{6}{p{0.9cm}}},
  cell{1}{1} = {r=3}{},
  cell{1}{2} = {c=6}{},  %
  cell{2}{2} = {c=2}{},  %
  cell{2}{4} = {c=2}{},
  cell{2}{6} = {c=2}{},
  hline{1,4,7} = {-}{0.12em},  %
  hline{2-3} = {-}{},          %
  column{2-7} = {m, colsep=2pt}, %
}
\textbf{Model} & \textbf{Task} & & & & & & \\
               & Pick \& Place & & Handover & & Scene Interact & & \\
               & \scriptsize{Plan $\uparrow$} & \scriptsize{Succ $\uparrow$} 
               & \scriptsize{Plan $\uparrow$} & \scriptsize{Succ $\uparrow$} 
               & \scriptsize{Plan $\uparrow$} & \scriptsize{Succ $\uparrow$} \\
Qwen2.5VL & 62\% & 44\% & 68\% & 60\% & 56\% & 36\% \\
VLP (D)     & 92\% & 72\% & 90\% & 84\% & 84\% & 72\% \\
VLP (V+D)   & 88\% & 68\% & 94\% & 84\% & 90\% & 80\% \\
\end{tblr}
\end{table}

We conducted real-world robot experiments to evaluate the planning feasibility and execution success rates of VLP models fine-tuned using two different strategies, compared with the original Qwen2.5-VL model, across multiple task scenarios. The results are summarized in Table \ref{tab:models}. We performed a total of 50 trials for three task categories, \texttt{pick \& place}, \texttt{handover object} and \texttt{scene interaction} (e.g., closing a drawer, disposing of trash). The results show that the fine-tuned models achieve a significantly higher planning feasibility rate (exceeding 90\%) compared to the original model, and both fine-tuned versions achieve execution success rates above 70\%.

\begin{figure}
 \centerline{\includegraphics[width=7cm]{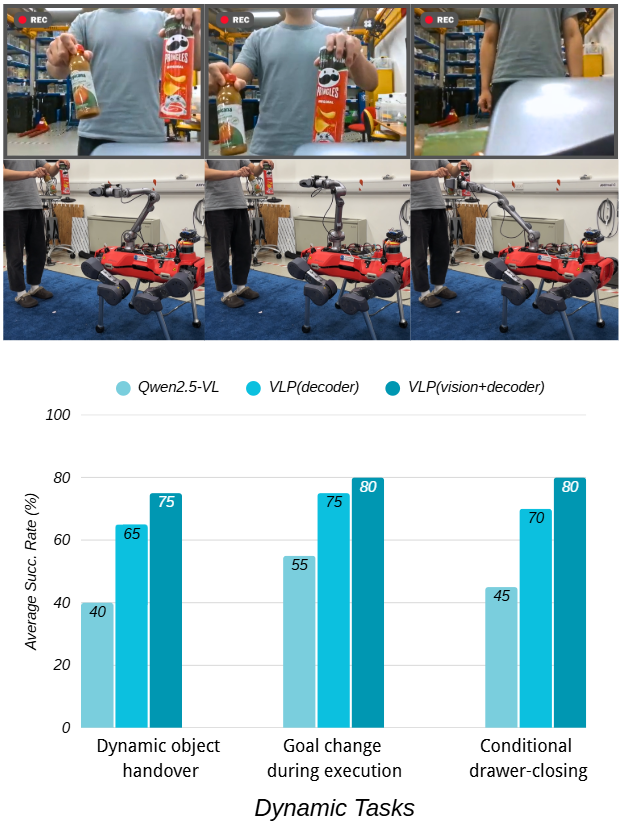}}
\caption{Top: Snapshots of a dynamic object handover task from different perspectives. Bottom: Model performance in dynamic task scenarios.}
\label{dynamic_task}   
\end{figure}

We further evaluated the model’s ability to dynamically update its strategy in response to changes in task requirements or scene state during execution through experiments conducted in dynamic environments. The \texttt{Dynamic\_object\_handover} task involves changing the handover object during execution; the \texttt{Goal\_change\_during\_execution} task requires modifying the target goal mid-process; and the \texttt{Conditional\_drawer\_closing} task requires the robot to monitor state conditions to complete the task. As shown in Fig. \ref{dynamic_task}, after 20 trials, the VLP model demonstrates substantially better performance in dynamic policy generation compared to the benchmark model, achieving over 20\% higher success rates. Moreover, due to requiring multiple rounds of scene reasoning and policy re-evaluation, the model fine-tuned on both the vision encoder and LM decoder achieves 5–10\% higher success rates than the decoder-only fine-tuned model.

\subsubsection{Cross-embodiment testing of the VLP model}

\begin{figure}
 \centerline{\includegraphics[width=7cm]{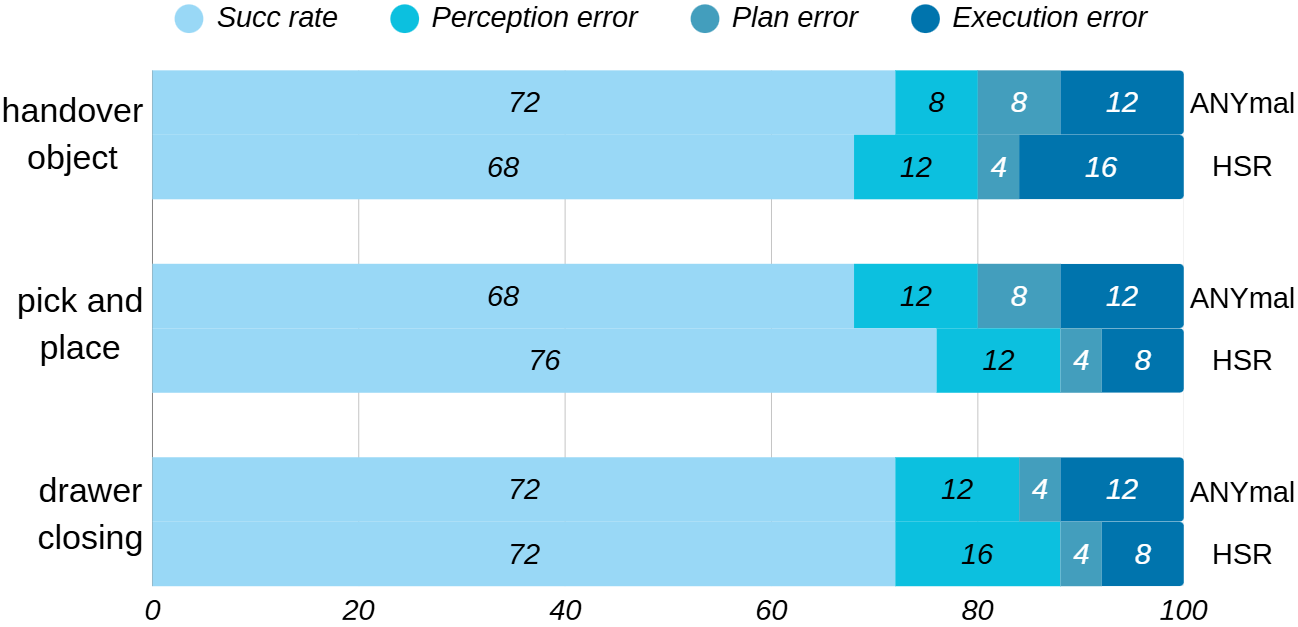}}
\caption{Evaluation of the models across different embodiments.}
\vspace{-1em}
\label{embodiment}   
\end{figure}

To evaluate the generalization capability and applicability of the proposed VLP model in task planning, we conducted experiments using different robot embodiments performing various manipulation tasks and recorded both task success rates and the proportion of different types of execution failures, as shown in Fig. \ref{embodiment}. The results indicate that across 25 trials for each of the three tasks on two different physical robot platforms, the model consistently achieved high task success rates (greater than 68\%). Most failures were attributed to perception or execution errors, which are highly dependent on the robot's hardware characteristics, while planning errors caused by the model accounted for less than 10\%. These results demonstrate the strong cross-embodiment adaptability of the proposed approach.

\subsection{Results Analysis}
The proposed VLP model demonstrates clear advantages over existing task planning methods, particularly in scenarios that require real-time re-planning. By integrating semantic reasoning with multimodal scene understanding and hierarchical policy representation, the system effectively adapts to instruction and environment changes during execution. Post-training on real-world interaction data leads to improved planning feasibility and execution stability relative to pre-trained baselines in our evaluation, facilitating online structured policy generation across multiple task categories. Furthermore, jointly fine-tuning both the visual encoder and LM decoder enhances scene reasoning and policy refinement, which proves especially beneficial under dynamic conditions. Finally, experiments across different robotic embodiments confirm the framework’s strong adaptability and generalization, showing its potential for deployment in practical robotic systems. Overall, the results validate that the VLP model supports interpretable planning, efficient policy updates, and rapid deployment, making it a promising solution for autonomous task execution in real-world environments.

\section{Conclusion}

In this work, we presented a VLP-based dynamic planning framework that unifies semantic understanding, multimodal reasoning, and hierarchical policy generation to enable autonomous task planning and execution across different robotic embodiments. By leveraging post-training on real-world interaction data and local inference, the proposed method achieves interpretable planning and real-time strategy adaptation in unstructured environments. Experimental results across multiple platforms and scenarios validate its effectiveness and generalization.

Despite these strengths, the current system is constrained by the predefined set of action and perception primitives, which may limit planning flexibility in open-set environment. Future work will focus on enhancing perception capabilities, integrating autonomous navigation and long-horizon planning, and expanding the skill library to support more complex tasks.



\bibliographystyle{IEEEtran}
\bibliography{references}

\addtolength{\textheight}{-12cm}   

\end{document}